\documentclass{article}
\usepackage{spconf,amsmath,graphicx}
\usepackage{booktabs}

\usepackage{epsfig}
\usepackage{graphicx}
\usepackage{amsmath}
\usepackage{amssymb}
\usepackage{gensymb}

\usepackage{xspace}
\makeatletter
\def\onedot{\ifx\@let@token.\else.\null\fi\xspace}
\makeatother
\def\ie{\textit{i.e}\onedot}

\newcommand{\Tref}[1]{Table~\ref{#1}}

\newcommand{\Fref}[1]{Figure~\ref{#1}}

\newcommand{\Cref}[1]{Chap.~\ref{#1}}

\newcommand{\calV}{{\mathcal{V}}}

\newcommand{\be}{\begin{eqnarray}}
\newcommand{\ee}{\end{eqnarray}}
\newcommand{\bee}{\begin{eqnarray*}}
\newcommand{\eee}{\end{eqnarray*}}

\newcommand{\matrixb}{\left[ \begin{array}}
\newcommand{\matrixe}{\end{array} \right]}

\newcommand{\app}{\raise.17ex\hbox{$\scriptstyle\sim$}}

\usepackage{enumitem}
\usepackage[normalem]{ulem}
\usepackage[ruled]{algorithm2e}
\usepackage{multirow}
\usepackage{array}
\usepackage{ctable} % for \specialrule command
\usepackage{makecell}
\usepackage[capposition=bottom]{floatrow}
\usepackage{comment}
\DeclareMathAlphabet{\mathcal}{OMS}{cmsy}{m}{n}
\usepackage{textcomp}
\usepackage{dblfloatfix}
\usepackage{caption}
\usepackage{subcaption}
\usepackage{colortbl,xcolor}
\usepackage{booktabs}
\usepackage{mathtools}
\usepackage{textpos}
\usepackage[flushmargin]{footmisc}

\title{LEARNING SOUND LOCALIZATION BETTER\\FROM SEMANTICALLY SIMILAR SAMPLES}
\name{Arda Senocak*, Hyeonggon Ryu*, Junsik Kim${}^\dagger$*, In So Kweon
\thanks{*Equal contributions are listed in alphabetical order. The part of this work was done when Junsik Kim was in KAIST and supported by Basic Science Research Program through the National Research Foundation of Korea(NRF) funded by the Ministry of Education(2021R1I1A1A01059778).}}
\address{KAIST, Daejeon, Korea\\
${}^\dagger$Harvard University, Cambridge MA, USA}
\begin{document}
%\ninept
%
\maketitle
\begin{abstract}
    The objective of this work is to localize the sound sources in visual scenes. Existing audio-visual works employ contrastive learning by assigning corresponding audio-visual pairs from the same source as positives while randomly mismatched pairs as negatives. However, these negative pairs may contain semantically matched audio-visual information. Thus, these semantically correlated pairs, ``hard positives'', are mistakenly grouped as negatives. Our key contribution is showing that hard positives can give similar response maps to the corresponding pairs. Our approach incorporates these hard positives by adding their response maps into a contrastive learning objective directly. We demonstrate the effectiveness of our approach on VGG-SS and SoundNet-Flickr test sets, showing favorable performance to the state-of-the-art methods. 
\end{abstract}
%
%-------------------------------------------------------------------------
\section{Introduction}\label{sec:intro}

During daily events of our lives, we are continuously exposed to various sensory signals and their interactions with each other. Because of this continuous stream of information, human perception has been developed to organize incoming signals, recognize the semantic information and understand the relationship between these cross-modal signals to combine or separate them. Among these sensory signals, inarguably the most dominant ones are vision and audition.  In order to have human-level perceptional understanding, modeling proper audio-visual learning that can associate or separate the audio-visual signals is essential. Thus, the audio-visual learning is an emerging research topic with variety of tasks, such as audio-visual source separation~\cite{ephrat2018looking, zhao2019sound, gao2019coSep, gan2020gestureSep, afouras2020AVObjects, gao2021visualVoice, tzinis2021audioScope}, audio spatialization~\cite{morgadoNIPS18, Yang_2020_CVPR, xu2021visually}, audio-visual video understanding~\cite{long2018pureAttention, korbar2019scsampler, kazakos2019epicFusion, xiao2020avSlowFast, gao2020listen2look, wang2020multimodalHard} and sound localization~\cite{arandjelovic2018objects, senocak2018learning, senocak2019learning, hu2020discriminative,  chen2021localizing, lin2021unsupervised}.

\begin{figure}[t]
    \centering
    \vspace{9mm}
    \resizebox{1\linewidth}{!}{
    \includegraphics{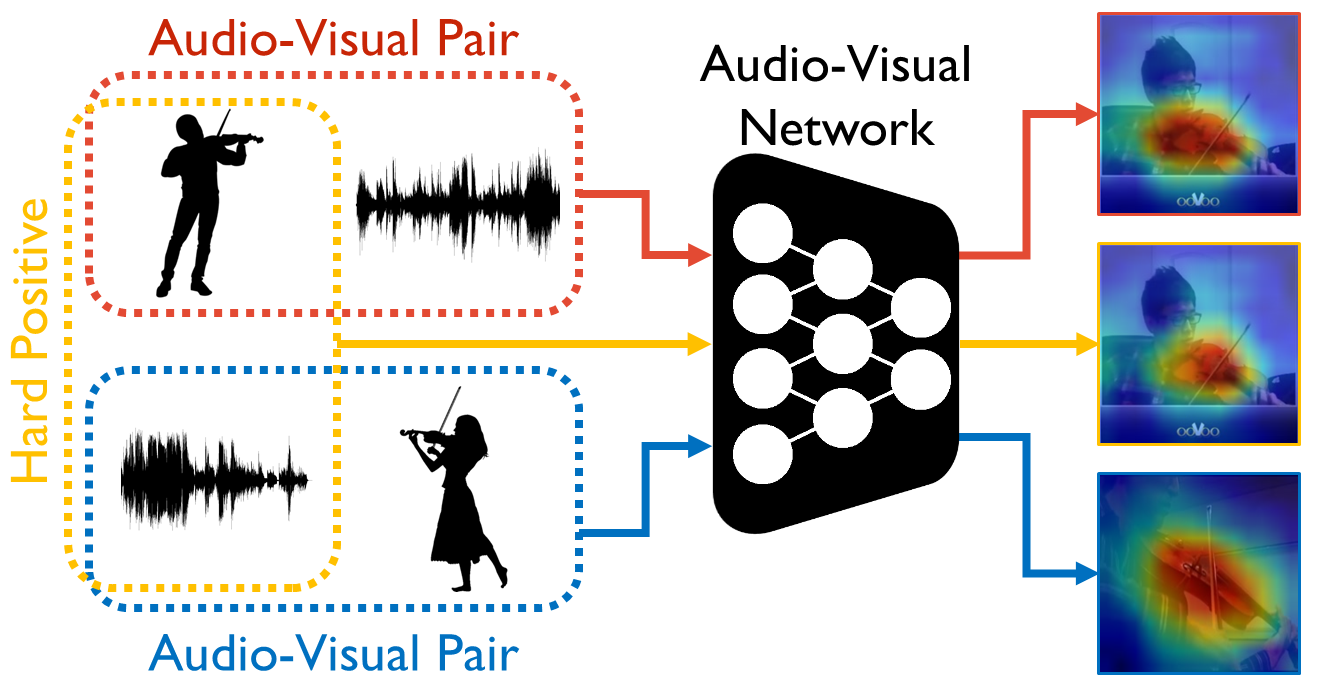}
    \caption{\textbf{Localizing Sound Sources by Introducing Hard Positive Samples.} 
    In this example, two audio-visual pairs from the red box and blue box are semantically related. When we pair the image of the red box and the audio of the blue box, which forms the yellow box, the response maps from the yellow box and the red box localize similar regions.
    }
    \label{fig:teaser}
    }
\end{figure}
One of the main challenging problems in audio-visual learning is to discover a sound source in a visual scene by taking advantage of the correlation between the visual and audio signals. A simple way to do this is using sounding source segmentation masks or bounding boxes as supervision. However, obtaining sufficient annotated data is difficult and costly for audio-visual tasks, where annotation often requires listening and watching samples. As a result, a key challenge is proposing unsupervised approaches, without any manual annotation, that can solve this task successfully. A widely used self-supervised approach in sound localization is using the correspondence between the audio and visual signals by using them as supervision to each other~\cite{arandjelovic2018objects, senocak2018learning, senocak2019learning, Owens2018AudioVisualSA}. Additionally,~\cite{senocak2018learning, senocak2019learning} use an attention mechanism to refine visual features by feature weighting with sound source probability map. Other models~\cite{hu2019deep,hu2020curriculum} deploy the clustering of audio-visual samples to reveal the cross-modal relationships and improve the accuracy. While prior self-supervised works ignore the category information,~\cite{hu2020discriminative} incorporates class-aware object dictionaries and distribution alignment in a self-supervised way. More recently, in the light of the success of the noise contrastive learning,~\cite{chen2021localizing} introduces a state-of-the-art method that uses a contrastive learning scheme with hard negative mining from background regions within an image.

Inarguably, audio-visual correspondence in a video, \ie audio and image pairs, is the key assumption in sound localization approaches. Vast amount of audio-visual research leverage contrastive learning~\cite{chopra2005learning, hoffer2015deep, infoNCE} by assigning corresponding audio-visual pair from the same source as positives while mismatched pairs, \ie audio and visual pairs from different sources, as negatives.
One problem of contrastive learning, when applied with this audio-visual correspondence assumption, is that the negative pairs may contain semantically matched audio-visual information. For example, if there are multiple videos with a person playing the violin in a dataset, every pairwise relation of these videos are semantically correct pairs (\Fref{fig:teaser}). However, when contrastive learning is applied without consideration of semantic information contained in videos, a model is falsely guided by these false negatives, \ie audio and video from different sources but containing similar semantic information~\cite{han2020coclr,morgado2021robust}.

In order to address the false negative issue in a sound localization task, we mine and incorporate these false negatives into the training. As a consequence, we can avoid using falsely paired audio-visual samples as negatives during the training. Moreover, since these pairs are semantically correlated, they can be used as positive pairs, \ie hard positives, in sound localization tasks.
For instance, if the audio samples of two different instances are semantically related, then the attention maps of those audios paired with the same base image should be similar to each other as in~\Fref{fig:teaser}. Note that this observation is valid when the order of the audio and vision samples in the previous scenario are swapped as well. We show this simple approach boosts sound localization performance on standard benchmarks. 

To be clear, we do not propose a new architecture or a new loss function in this paper, but instead, we provide a new training mechanism by discovering semantically matched audio-visual pairs and incorporating them as hard positives. We make the following contributions: 1) We demonstrate hard positive audio-visual pairs produce similar localization results to that of corresponding pairs. 2) We mine and incorporate hard positives into the positive set of contrastive loss. 3) We show that removing hard positives from the negative set improves the sound localization performance and outperforms prior works on the standard benchmarks.

%-------------------------------------------------------------------------
\vspace{-5mm}\section{Approach}\label{sec:approach}
Audio-visual attention is commonly used in sound localization studies~\cite{senocak2018learning, hu2020discriminative, chen2021localizing}. We build the baseline audio-visual model based on the most recent work LVS~\cite{chen2021localizing} and validate our method on top of the baseline. We first introduce the audio-visual attention mechanism and the baseline model named vanilla-LVS, and then introduce our approach.

\subsection{Preliminaries}

We use an image frame $v\in \mathbb{R}^{3\times H_v \times W_v}$ and the corresponding spectrogram $a\in\mathbb{R}^{1\times H_a \times W_a}$ of the audio from a clip $X=\{v, a\}$. With the two-stream  models, $f_v(\cdot; \theta_v)$ for vision embedding and $f_a(\cdot; \theta_a)$ for audio embedding, the signals are encoded into the features:
\begin{equation}
\begin{aligned}
V &= f_v(v; \theta_v), \quad V\in\mathbb{R}^{c\times h\times w}
\\
A &= f_a(a; \theta_a), \quad A\in\mathbb{R}^{c}
\end{aligned}
\end{equation}
The vision and audio features, $V_j$ and $A_i$, are fed into the localization module and audio-visual response map $\alpha_{ij}\in~\mathbb{R}^{h~\times w}$ is computed with cosine similarity as:
\begin{align}
[\alpha_{ij}]_{uv} =
\frac
{\langle A_i, [V_j]_{:uv}\rangle}
{\|A_i\|~\|[V_j]_{:uv}\|},
\quad
uv \in [h] \times [w],
\end{align}
where $i$ and $j$ denote the audio and image sample indices respectively. 
Following~\cite{chen2021localizing}, pseudo-ground-truth mask ${m}_{ij}$, is obtained by thresholding the response map as follows: 
\begin{align}
{m}_{ij} &= \sigma((\alpha_{ij}- \epsilon)/ \tau),
\end{align}
where $\sigma$ refers to the sigmoid function, $\epsilon$ to the thresholding parameter and $\tau$ is the temperature. The inner product between the mask $m_{ij}$ and the response map $\alpha_{ij}$ is computed to emphasize positively correlated regions of the response map.

\vspace{-4mm}\subsection{Semantically Similar Sample Mining}\label{sec:pos_mine}
Our method is based on the observation that hard positives make similar localization results to the original pairs. These additional audio-visual response maps from the hard positives can be easily incorporated into the contrastive learning formulation. Semantically related samples in each modality are mined to form hard positives based on the similarity scores in the feature space. To get the reliable representations within each modality, we train the baseline model without 
tri-map, \ie hard negative mining introduced in ~\cite{chen2021localizing}.

Given sample $i$ and an arbitrary sample $j$, we compute the cosine similarity between the features within each modality, $A_i^T  A_j$ and $\calV_i^T  \calV_j$, where $\calV_i$ is the spatial-wise average pooled vector of visual the feature map $V_i$. 
Sets of semantically similar items in both modality for the given sample, $\mathcal{P}_{i_A}$ and $\mathcal{P}_{i_V}$, are constructed based on the computed scores, $\mathcal{S}_{i_A}$ and $\mathcal{S}_{i_V}$.~$K$ number of semantically related samples are retrieved from the sorted similarity scores. All the samples that are not contained in the aforementioned sets are considered as negative samples and form a negative set, $\mathcal{N}_{i}$ :

\begin{equation}
\begin{aligned}
\mathcal{S}_{i_A} &= \{ A_i^T  A_j | 1 \leq j \leq n \}, 
\\
\mathcal{P}_{i_A} &= \{X_t\}_{t\in S[1:K]}, \quad S=argsort({S_{i_A}}),  
\\
\mathcal{S}_{i_V} &= \{ \calV_i^T  \calV_j | 1 \leq j \leq n \}, 
\\
\mathcal{P}_{i_V} &= \{X_t\}_{t\in S[1:K]}, \quad S=argsort({S_{i_V}}),  
\\
\mathcal{N}_{i} &= \overline{{\mathcal{P}}_{i_A} \cup {\mathcal{P}}_{i_V}}.
\end{aligned}
\end{equation}

\subsection{Training}\label{sec:training}
The training objective of our method makes positive pair responses higher while negative pair responses are reduced. Since we incorporate responses from hard positives, we extend the contrastive learning formulation of~\cite{chen2021localizing} by adding each hard positive response. Defining the response map of the base pairs as ${P}_b$, hard positive responses are computed as follows: ${P}_a$ is the response of base visual signal and semantically similar audio, ${P}_v$ is the response of base audio signal and semantically similar image, finally ${P}_c$ is the cross-modal hard positive pair\textquotesingle s response.  
All the responses from the base audio and negatively correlated image pairs are considered as the negative responses ${N}_{i}$. Definition of positive and negative responses are given as: 

\begin{equation}
\begin{aligned}
P_b &= \frac{1}{|m_{ii}|}
\langle  m_{ii}, ~\alpha_{ii} \rangle
\\
P_a &= \frac{1}{|m_{ji}|}
\langle  m_{ji}, ~\alpha_{ji} \rangle , j \in \mathcal{P}_{i_A}
\\
P_v &= \frac{1}{|m_{ik}|} 
\langle  m_{ik}, ~\alpha_{ik} \rangle , k \in \mathcal{P}_{i_V}
\\
P_c &= \frac{1}{|m_{jk}|} 
\langle  m_{jk}, ~\alpha_{jk} \rangle , j \in \mathcal{P}_{i_A}, k \in \mathcal{P}_{i_V}
\\
P_i &= \exp(P_b) + \exp(P_a) + \exp(P_v) + \exp(P_c)
\\
N_i &=
\sum_{l \in \mathcal{N}_i} \exp(\frac{1}{hw} 
\langle \mathbf{1}, ~\alpha_{il}\rangle).
\end{aligned}
\end{equation}
After constructing positive and negative responses, our model can be optimized by the loss function $\mathcal{L}$ as below: 
\begin{align}
\mathcal{L} &= 
% -\mathbb{E}_{d_i \sim \mathcal{D}} 
-\frac{1}{n}\sum_{i=1}^n
\left[ \log \frac{P_i}{P_i + N_i} \right]
\end{align}

\begin{table}[ht]
\centering
% \small
% \vspace{1em}
\begin{tabular}{ccccc}
\toprule
& \multicolumn{2}{c}{VGG-SS} & \multicolumn{2}{c}{Flickr} \\
Method & cIoU & AUC & cIoU & AUC \\ \midrule
Attention~\cite{senocak2018learning}  &  0.185 & 0.302 & 0.660 & 0.558 \\
AVEL~\cite{tian2018audio}     	       & 0.291 & 0.348 & - & - \\
AVObject~\cite{afouras2020AVObjects}  	 & 0.297 & 0.357 & - & - \\
Vanilla-LVS       		& 0.278 & 0.350 & 0.692 & 0.563 \\
LVS~\cite{chen2021localizing}$\dagger$    & 0.303 & 0.364 & 0.724 & 0.578 \\
Random HP   		       & 0.207 & 0.314  & 0.572 & 0.528 \\
\textbf{Ours}    & \textbf{0.346} & \textbf{0.380} & \textbf{0.768} & \textbf{0.592} \\ \bottomrule
\end{tabular}%
{\vspace{-3pt}
\caption{\textbf{Quantitative results on the VGG-SS and SoundNet-Flickr test sets}. All models are trained with 144K samples from VGG-Sound and tested on VGG-SS and SoundNet-Flickr. $\dagger$ is the result of the model released on the official project page and the authors report ~3$\%$ drop in cIoU performance comparing to their paper.}\label{tab:quantitative}}
\end{table}
%-------------------------------------------------------------------------
\section{Experiments}\label{sec:experiments}
% \vspace{-4mm}

\begin{table}[t]
\centering
% \small
% \vspace{1em}
\begin{tabular}{lcc}
\toprule
Method        		    & cIoU    & AUC \\ \midrule
Attention\cite{senocak2018learning} & 0.660 & 0.558 \\
Vanilla-LVS       		& 0.704 & 0.581 \\
LVS~\cite{chen2021localizing}$\dagger$  	       & 0.672 & 0.562 \\
\textbf{Ours}         & \textbf{0.752} & \textbf{0.597} \\
\bottomrule
\end{tabular}%
{\vspace{-3pt}
\caption{\textbf{Quantitative results on the SoundNet-Flickr test set.} All models are trained and tested on the SoundNet-Flickr dataset. $\dagger$ is the result of the model from the official project page.}\label{tab:quantitative_second}}
\end{table}

\begin{figure}[t]
    \centering
    \resizebox{1\linewidth}{!}{
    \includegraphics{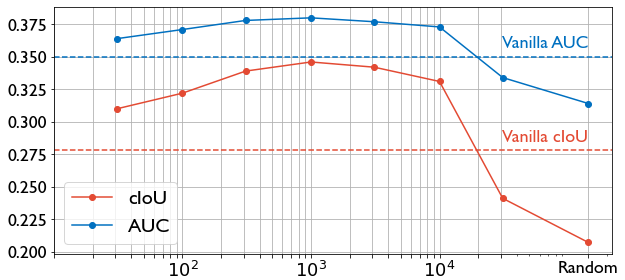}
    \caption{\textbf{Localization performance with varying $K$ values}.\vspace{-2mm}}
    \label{fig:teaser_hyeonggon}
    }
    \vspace{-4mm}
\end{figure}

\begin{figure*}[t]
\centering
{
\resizebox{1\linewidth}{!}{%
\begin{tabular}{c}
\includegraphics[width = 1\linewidth]{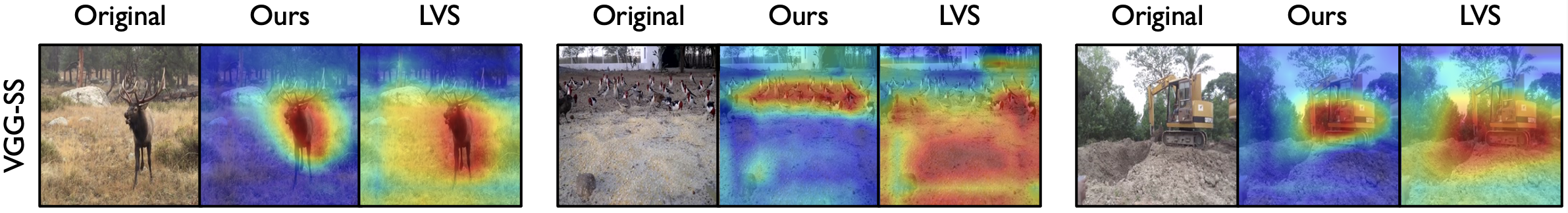} \\
\includegraphics[width = 1\linewidth]{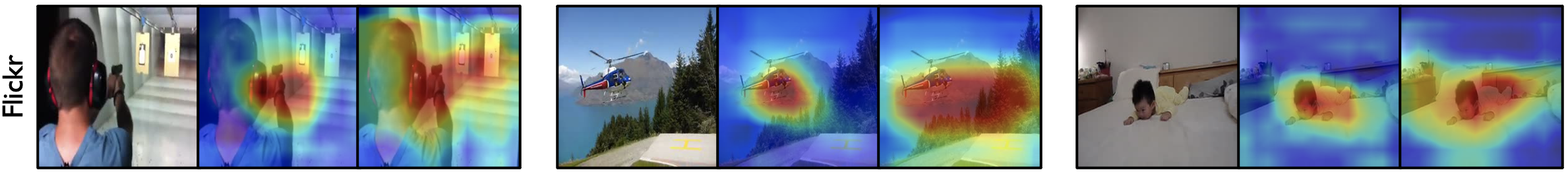} \\
\end{tabular}
}
}
\caption{\textbf{Sound localization results on VGG-SS and SoundNet-Flickr and comparison with LVS~\cite{chen2021localizing}.}}
\label{fig:qualitative_vggss}
\vspace{-4mm}
\end{figure*}

\begin{figure*}[t]
\centering
{
\resizebox{1\linewidth}{!}{%
\begin{tabular}{c}
\includegraphics[width = 1\linewidth]{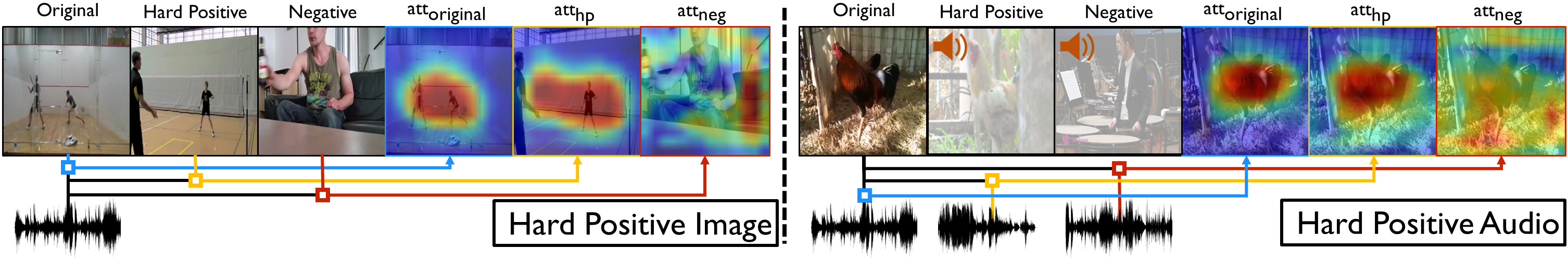} \\
\end{tabular}
}
}
\caption{\textbf{Response maps of hard positives.}  Left refers to the scenario that the hard positive is obtained with the visual signal. Right shows the hard positive pair which is obtained with the audio signal. In both cases, the response maps of hard positives resemble the original pair response maps.}
\label{fig:qualitative_hps}
\vspace{-4mm}
\end{figure*}

\vspace{-2mm}
\subsection{Datasets}\label{sec:datasets}
We train our method on VGGSound~\cite{VGGSound} and SoundNet-Flickr~\cite{aytar2016soundnet} and test with VGG-SS~\cite{chen2021localizing} and SoundNet-Flickr-Test~\cite{senocak2018learning} datasets. VGGSound is a recently released audio-visual dataset containing around ~200K videos. SoundNet-Flickr-training set is provided by~\cite{senocak2018learning,senocak2019learning}. Test splits of VGG-SS and SoundNet-Flickr-Test, have bounding box annotations of sound sources for ~5K and 250 samples respectively.

\vspace{-2mm}
\subsection{Implementation Details}\label{sec:implementation}
VGGSound dataset contains video clips. We use the center frames of the video clips for the training. In the SoundNet-Flickr dataset, still images are paired with the audio samples. 
The $257 \times 300$ magnitude spectrogram from a 3s audio segment around the frame is used as the corresponding audio signal. The ResNet18~\cite{resnet} is used as a backbone for each modality. Following~\cite{chen2021localizing}, we use $\epsilon =0.65$, $\tau = 0.03$. $K=1000$ is used to mine top-$K$ semantically similar samples.

\vspace{-2mm}
\subsection{Quantitative Results}\label{sec:quantitative}
In this section, we compare our results with existing sound localization approaches on VGG-SS and SoundNet-Flickr-Test datasets. We recall that our model is based on vanilla-LVS trained with semantically similar sample mining.
In ~\Tref{tab:quantitative}, we show the performance of the proposed model together with several prior works on VGG-SS and SoundNet-Flickr-Test datasets by following~\cite{chen2021localizing}. The comparison methods here are trained with the same amount of training data, 144K, as in~\cite{chen2021localizing}. AVEL~\cite{tian2018audio} and AVObject~\cite{afouras2020AVObjects} models are based on video input. Thus, the SoundNet-Flickr dataset, which contains static image and audio pairs, can not be evaluated. 
The proposed model achieves significantly higher performance on both the VGG-SS and SoundNet-Flickr-Test datasets than the other existing works including LVS. This shows that inter-sample relation across-modality is more important than the intra-sample relation.

Next, we compare the performance on the SoundNet-Flickr-Test set by training our method separately with 144K samples from SoundNet-Flickr. As shown in~\Tref{tab:quantitative_second}, the proposed model gives the highest performance in this comparison. As~\cite{chen2021localizing} reports, our method also achieves higher accuracy on this test set when it is trained with VGGSound.

%Ablation Study
In order to see the effect of the value of $K$ used to select the $K$ semantically related samples, we report ablative evaluation in~\Fref{fig:teaser_hyeonggon} with the model trained on VGGSound. 
Our model gains performance over the baseline in a broad range of value $K$ showing that the proposed method is beneficial without careful hyperparameter tuning.
Results show that a choice of $K = 1000$ is optimal. This can be explained from the perspective of the sample distribution of VGGSound~\cite{VGGSound} as each class contains 300–1000 clips. Considering the number of the semantically similar classes and the number of their samples for the given instance, our reported numbers show an expected trend from $K = 300$ to $K = 3000$. We also report the result of our method by constructing a positive set with randomly selected samples. The low performance here shows that randomly forming hard positives hurts the training as it brings extreme noise, \ie semantically mismatched pairs.

\subsection{Qualitative Results}\label{sec:qualitative}
We provide our sound localization results on VGG-SS and SoundNet-Flickr test samples in~\Fref{fig:qualitative_vggss} and compare them with LVS~\cite{chen2021localizing}. 
Our results present more accurate response maps in comparison to the competing approach. Additionally, we demonstrate attention maps of the hard positives. The left part of the~\Fref{fig:qualitative_hps} visualizes the scenario where hard positive is obtained with the visual signal. Here, the original audio is paired with the semantically related image. Similarly, the right part of the~\Fref{fig:qualitative_hps} depicts that the hard positive is obtained with the audio signal. Results show that in both scenarios, response maps of the hard positives are very similar to the response maps of the original pairs, ${att}_{original}$. As expected, response maps of the negative samples, ${att}_{neg}$, are inaccurate since those instances are not correspondent.

%-------------------------------------------------------------------------
% \vspace{-4mm}
\section{Conclusion}
In this paper, we address the problem of self-supervised sound source localization in contrastive learning formulation. We identify a source of noise due to the random negative sampling, where semantically correlated pairs are contained among negative samples. We suggest a simple positive mining approach and employ these hard positives into training. We validate our method on standard benchmarks showing state-of-the-art performance. The proposed method is applicable to any type of model using contrastive loss, therefore audio-visual learning studies can benefit from our work.

% \clearpage
%-------------------------------------------------------------------------

%\vfill
%\pagebreak

\newpage
\bibliographystyle{IEEEbib}
\ninept
\bibliography{strings,refs}

\end{document}